\pdfoutput=1

\documentclass[11pt]{article}

\usepackage[]{acl}

\usepackage{times}
\usepackage{latexsym}
\usepackage{comment}
\usepackage[T1]{fontenc}

\usepackage[utf8]{inputenc}
\usepackage{microtype}

\usepackage{amsmath,amsfonts,bm}

\def\eqref#1{equation~\ref{#1}}

\def\1{\bm{1}}

\DeclareMathAlphabet{\mathsfit}{\encodingdefault}{\sfdefault}{m}{sl}
\SetMathAlphabet{\mathsfit}{bold}{\encodingdefault}{\sfdefault}{bx}{n}

\usepackage{hyperref}
\usepackage{url}

\usepackage{amsfonts}
\usepackage[ruled]{algorithm2e}

\usepackage{enumitem}
\usepackage{pgfplots}
\usepackage{arydshln}
\usepackage{pifont}
\usepackage{amsmath,amssymb}
\usepackage{multicol,multirow}
\usepackage{amssymb}
\pgfplotsset{width=7cm,compat=1.13}
\usepackage{microtype}
\usepackage{graphicx}
\usepackage{subfigure}
\usepackage{times}
\usepackage{latexsym}
\usepackage{booktabs} 
\usepackage{bbm}

\usepackage{tikz,pgfplots}
\usepgfplotslibrary{groupplots}
\usepackage{nicefrac} 
\pgfplotsset{compat=1.13}

\definecolor{g-red}{HTML}{DB4437}
\definecolor{g-blue}{HTML}{4285F4}
\definecolor{g-green}{HTML}{0F9D58}
\definecolor{g-yellow}{HTML}{F4B400}
\definecolor{g-orange}{HTML}{FF9800}
\definecolor{g-grey}{HTML}{9E9E9E}
\definecolor{shannon}{HTML}{304FFE}
\definecolor{uw}{RGB}{138,43,226}
\definecolor{stanford}{RGB}{255,69,0}
\definecolor{const}{RGB}{68, 110, 182}
\definecolor{head}{RGB}{246, 180, 32}
\definecolor{freq}{RGB}{0, 0, 0}
\definecolor{ao}{rgb}{0.0, 0.5, 0.0}
\definecolor{asparagus}{rgb}{0.53, 0.66, 0.42}
\definecolor{amber}{rgb}{1.0, 0.49, 0.0}
\definecolor{alizarin}{rgb}{0.82, 0.1, 0.26}
\definecolor{applegreen}{rgb}{0.55, 0.71, 0.0}
\definecolor{amethyst}{rgb}{0.6, 0.4, 0.8}
\definecolor{auburn}{rgb}{0.43, 0.21, 0.1}

\newcommand{\todo}[1]{\textcolor{magenta}{\bf\small [todo - #1]}}
\newenvironment{tightitemize}%
  {\begin{itemize}[topsep=0.5pt, partopsep=0.5pt] %
    \setlength{\itemsep}{0.5pt}%
    \setlength{\parskip}{0.5pt}%
    }%
  {\end{itemize}}

\usepackage{caption}
\usepackage{amssymb}

\newcommand{\heart}{\text{\small \ding{170}}}
\usepackage{babel}

\renewenvironment{quote}{%
  \list{}{%
    \leftmargin0.3cm   
    \rightmargin\leftmargin
  }
  \item\relax
}
{\endlist}
\title{Text Classification via Large Language Models}

\author{
Xiaofei Sun$^{\blacklozenge}$*, Xiaoya Li$^{\clubsuit}$*, 
Jiwei Li$^{\blacklozenge}$, Fei Wu$^{\blacklozenge}$ \\
{\bf Shangwei Guo$^{\blacktriangle}$, Tianwei Zhang$^{\heart}$, Guoyin Wang$^{\bigstar}$}}

\begin{document}
\maketitle
\begin{abstract}
Despite the remarkable success of
large-scale Language Models (LLMs) such as GPT-3,
their performances  still significantly underperform 
fine-tuned
models  in the task of text classification.
This is 
due to 
(1) the lack of reasoning ability in addressing complex  linguistic phenomena (e.g., intensification, contrast, irony etc);
(2)
limited number of tokens 
allowed in in-context learning. 

In this paper, we introduce \textbf{C}lue \textbf{A}nd \textbf{R}easoning \textbf{P}rompting (CARP).
CARP adopts a progressive reasoning strategy tailored to addressing the complex linguistic phenomena involved in text classification:
CARP
first prompts LLMs to find superficial clues 
   (e.g., keywords, tones, semantic relations, references, etc),
based on which  a diagnostic reasoning process is induced for final decisions. 
To further address the limited-token issue, 
CARP
 uses a fine-tuned model on the supervised dataset for $k$NN demonstration search in the in-context learning,
allowing the model to take the advantage of both 
LLM's generalization ability and the task-specific evidence provided by the  full labeled dataset.

Remarkably,
CARP yields new SOTA performances on 4
out of  5 widely-used text-classification benchmarks,  97.39 (+1.24) on SST-2,  96.40 (+0.72) on AGNews, 98.78 (+0.25) on R8 and 96.95 (+0.6) on R52,
and a performance comparable to SOTA on MR (92.39 v.s. 93.3). 
More importantly, we find that CARP delivers impressive abilities 
on low-resource 
and domain-adaptation
 setups. 
Specifically, 
using 16 examples per class, CARP achieves comparable performances to supervised models with 1,024 examples per class. Code and data are available at \href{https://github.com/ShannonAI/GPT-CLS-CARP}{github.com/ShannonAI/GPT-CLS-CARP}
\footnote{~* indicates equal contributions.}
\footnote{~$^{\blacklozenge}$Zhejiang University, $^{\clubsuit}$ Shannon.AI, $^{\bigstar}$Amazon \newline 
$^{\heart}$Nanyang Technological University, 
$^{\blacktriangle}$ Chongqing University \newline 
\{xiaofei\_sun, wufei, jiwei\_li\}@zju.edu.cn \newline 
xiaoya\_li@shannonai.com, swguo@cqu.edu.cn\newline 
tianwei.zhang@ntu.edu.sg, 
 guoyiwan@amazon.com}
\end{abstract}

\section{Introduction}
Large language models (LLMs)~\citep{gpt1, xue2020mt5, zhang2022opt, rae2021scaling, brown2020language, chowdhery2022palm, ouyang2022training, thoppilan2022lamda}
have shown the ability for in-context learning (ICL). Given a few demonstration examples, LLMs are prompted to generate results for a new test example, and have achieved performance comparable to supervised baselines or even state-of-the-art results in a variety of natural language processing (NLP) tasks such as question answering~\citep{trivedi2022interleaving}, natural language inference,~\citep{schick2020exploiting},
named entity recognition \cite{wang2023gpt}, relation extraction \cite{wan2023gpt}
 and information extraction~\citep{han2021ptr}.

\begin{figure*}[h]
 \centering
 \begin{minipage}[c]{\textwidth}
 \centering
\includegraphics[scale=0.5]{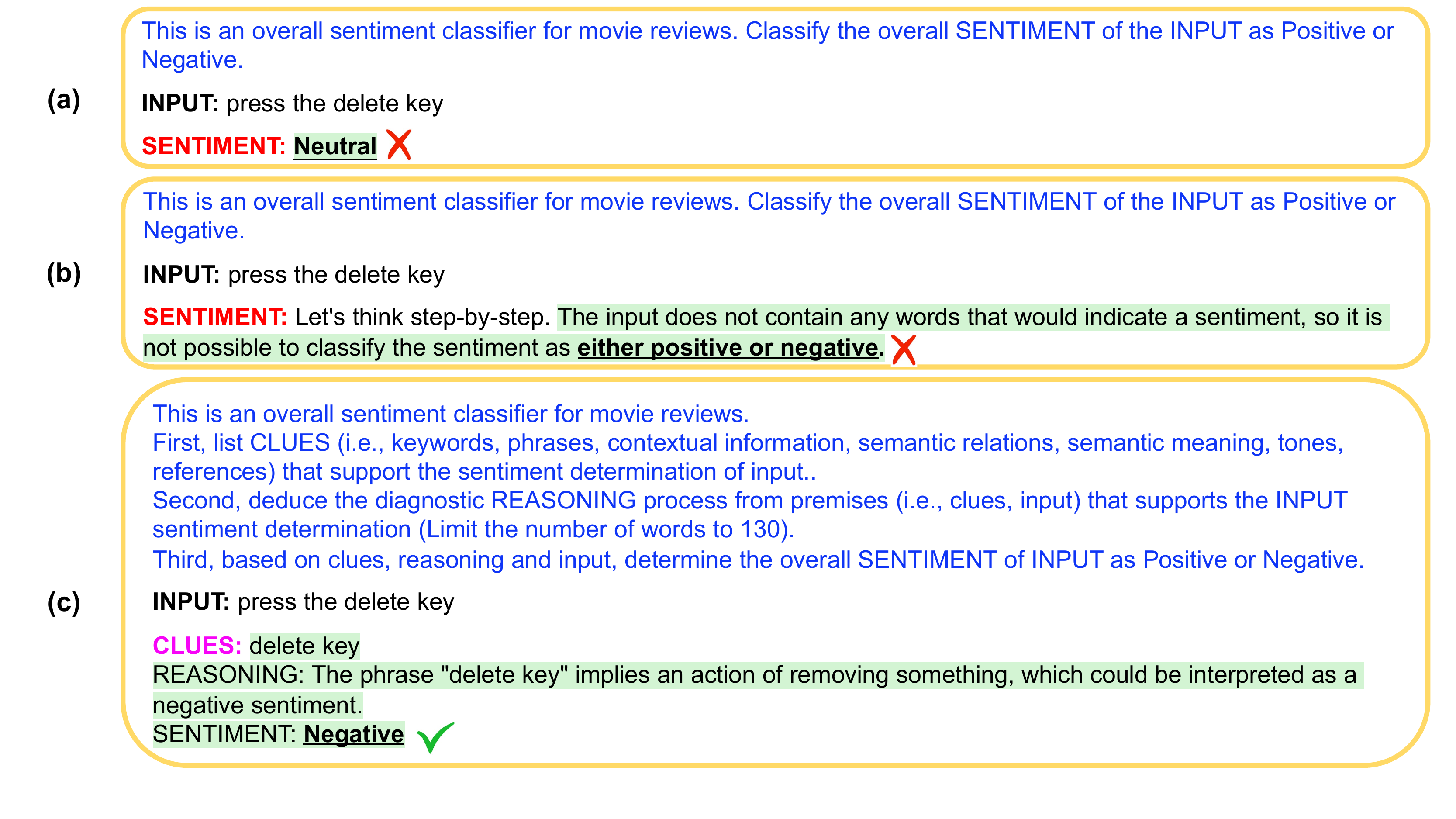}
\caption{Examples of zero-shot prompting methods for the text classification task: \textbf{(a)} represents for the \textbf{vanilla} prompting method; \textbf{(b)} denotes for the \textbf{Chain-of-Thought (CoT)}~\citep{Kojima2022LargeLM} prompting method; \textbf{c} represents for the proposed \textbf{CARP} prompting method.}
\label{fig:sample_figure}
 \end{minipage}
 \end{figure*}

\begin{figure*}[h]
 \centering
 \begin{minipage}[c]{\textwidth}
 \centering
\includegraphics[scale=0.45]{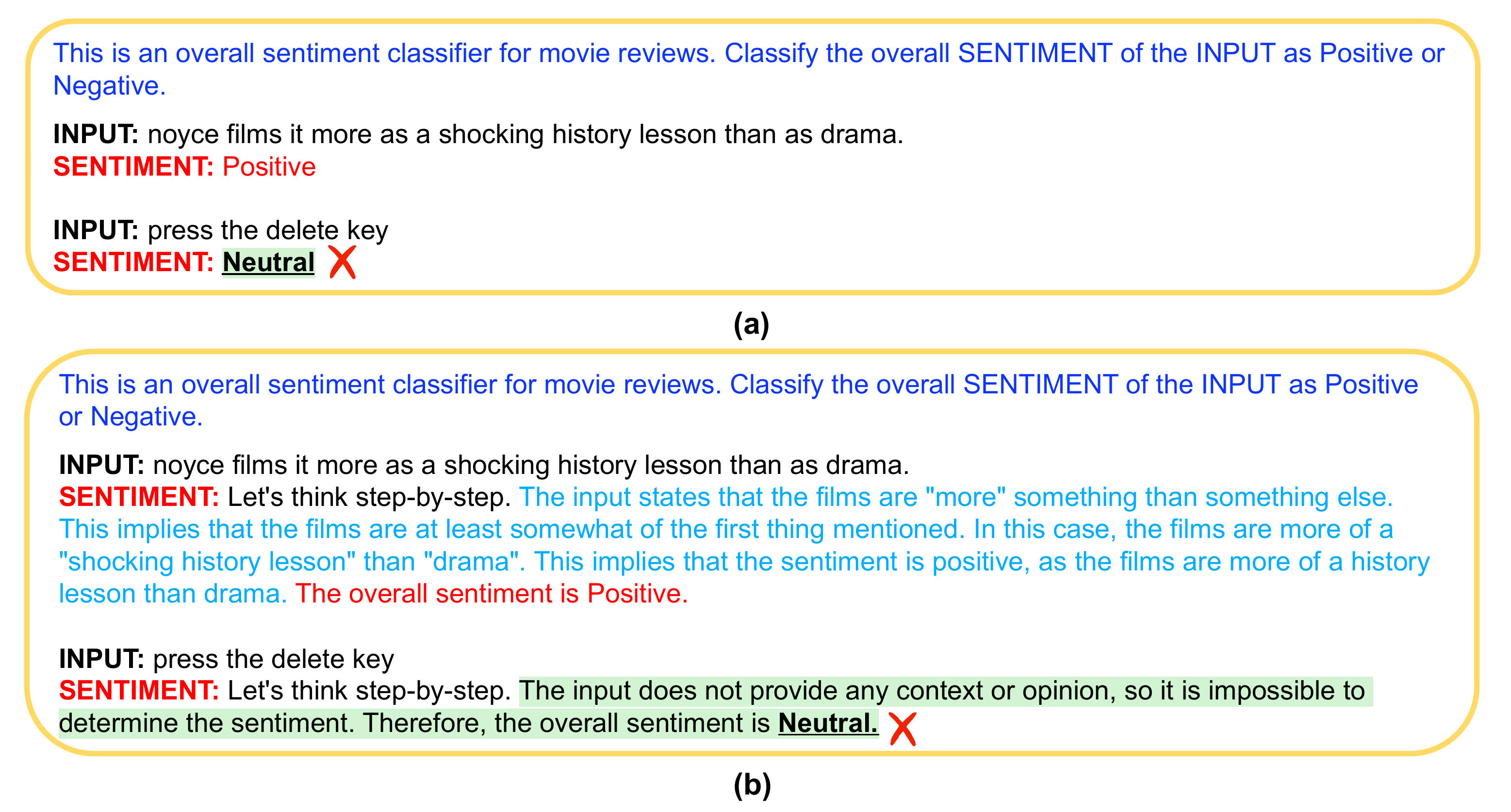}
\includegraphics[scale=0.45]{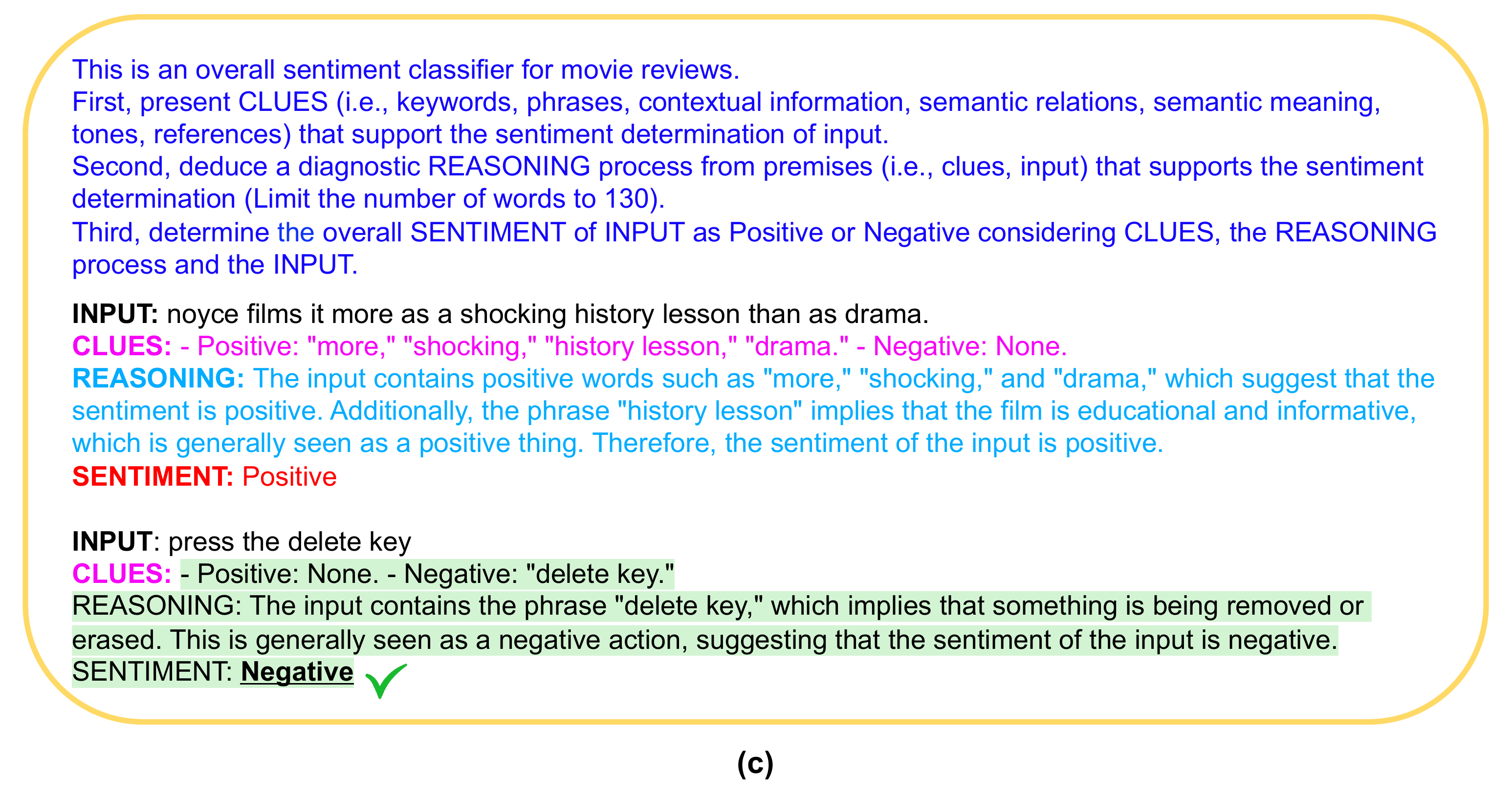}
\end{minipage}
\caption{Examples of few-shot ($k$=1) prompting methods for the text classification task: \textbf{(a)} represents for the \textbf{vanilla} prompting method; \textbf{(b)} denotes for the \textbf{Chain-of-Thought (CoT)}~\citep{Kojima2022LargeLM} prompting method; \textbf{(c)} represents for the proposed \textbf{CARP} prompting method.} 
\label{fig:few_shot}
\end{figure*}

In spite of the success, LLMs with ICL still significantly underperform
fine-tuned models  for text classification.
This is
 due to two reasons:
(1) Text classification 
requires models with more powerful reasoning abilities to 
resolve complex linguistic phenomenon 
including clause composition (e.g., concession, 
 negation, intensification), irony, etc. 
Recent efforts to improve LLMs' reasoning capabilities \citep{wei2022chain, Kojima2022LargeLM, ye2022unreliability, zhang2022automatic} mainly focus on tackling math problems, and thus are not tailored to addressing the  reasoning process necessary for the multitude of intricate linguistic phenomena in text classification;
(2) This  number of demonstration examples allowed in in-context learning is limited, e.g., the longest context allowed for GPT-3 is 4,096 subtokens. 
Therefore, LLMs are only able to take the advantage of a small proportion of the training set, performing well below supervised baselines;

In this paper, we introduce \textbf{C}lue \textbf{A}nd \textbf{R}easoning \textbf{P}rompting (CARP),  an extensible, annotation-free and efficient 
framework for text classification via large language models. 
To address 
 the  reasoning process necessary for handling the 
  linguistic phenomena in 
  text classification, 
  CARP decomposes the reasoning process into three steps, where LLMs are first prompted to find superficial clues 
   (e.g., keywords, tones, semantic relations, etc) in the given text; 
next, CARP treats 
 the clues and input as premises and
induce a diagnostic reasoning process; and 
finally determine the final label considering the above two steps. 
We find this progressive reasoning strategy to be effective in enhancing LLMs' ability in language reasoning involved in text classification.
Due to the limited number of tokens allowed in context, a more effective demonstration search is needed. 
CARP uses a fine-tuned model on the supervised dataset for $k$NN demonstration search for in-context learning. 
    Since the fine-tuned model is trained based on task-specific labels, 
    it guarantees that retrieved samples are close to the input sequence with respect to the task. 
Using fine-tuned models for demonstration search provides a channel to connect LLMs with the full training set, in spite of the limited number of tokens allowed in demonstrations. This strategy lets the model take the
advantage of both the LLMs' generalization abilities and all task-specific evidence provided by the training dataset.

Remarkably,
CARP yields new SOTA performances on four
out of  5 widely-used text-classification benchmarks,  97.39 (+1.24) on SST-2,  96.40 (+0.72) on AGNews, 98.78 (+0.25) on R8 and 96.95 (+0.6) on R52,
and a performance comparable to SOTA on MR (92.39 v.s. 93.3).
More importantly, we find that CARP delivers impressive ability 
on low-resource 
and domain adaptation
 setups with orders of magnitude fewer training examples. Specifically, 
CARP achieves comparable performances
with 16 examples per class 
 to supervised models trained on the full training set containing more than 1 thousand examples per class.
This demonstrates the capabilities of CARP 
in real-world text classification cases where  training data is limited.

\section{Related Work}
\subsection{Large Language Models}
Large language models (LLMs) are models that are trained using self-teaching algorithms on large unlabeled corpora. 
With emergent capabilities~\citep{xie2021explanation, wei2022emergent}, LLMs achieve significant performance boosts in NLP tasks. 

LLMs can be broadly divided into three categories based on the model architecture. 
The first category is the encoder-only model like BERT~\citep{devlin2018bert}.
BERT (300M)~\citep{devlin2018bert} and its variants~\citep{liu2019roberta, sun2020ernie, clark2020electra, feng2020codebert, sun2021chinesebert}  
adopt the {\it pre-training then fine-tuning} paradigm for NLP tasks: 
use masked language models as the main training objective for pretraining, and fine-tune the pretrained model in the annotated downstream datasets. 

The second category is the decoder-only models like GPT~\citep{gpt1}.
GPT~\citep{gpt1} uses the decoder of an auto-regressive transformer~\citep{vaswani2017attention} model for predicting the next token in a sequence.
GPT~\citep{gpt1} and its variants~\citep{dai2019transformer, keskar2019ctrl, radford2019language, chowdhery2022palm, zhang2022opt} also follow the {\it pre-training then fine-tuning} paradigm. 
GPT-3 (175B)~\citep{brown2020language} proposes to formalize all NLP tasks as generating textual responses condition on the given prompt. 

The third category is the encoder-decoder models like T5~\citep{raffel2020exploring}.
T5 (11B)~\citep{raffel2020exploring} and its variants~\citep{lewis2019bart, xue2020mt5} are encoder-decoder transformer models, which generate new sentences depending on a given input, following the  {\it pre-training then fine-tuning} paradigm.

\subsection{In-context Learning}
Unlike the {\it pre-training then fine-tuning} paradigm~\citep{devlin2018bert}, which saves model weights and uses task-specific datasets (i.e., train/valid/test set), in-context learning (ICL) generates textual responses (i.e., label words) conditioning on the given prompt (usually) with a few annotated examples for downstream tasks.

\newcite{li2021prefix, zhong2021factual, qin2021learning} propose to optimize prompts in the continuous space. 
\newcite{rubin2021learning, das2021case, liu2021makes, su2022selective} introduce different strategies for selecting in-context examples. 
\citet{lampinen2022can} show that explanations of examples in a few-shot prompt lead to a performance boost.
\citet{marasovic2021few} find that GPT-3 outperforms other models by a large margin in the explanation generation task.
\citet{wei2022chain} propose chain-of-thought reasoning and utilized <input, chain-of-thought, output> triples as the prompt for LLMs. \citet{wiegreffe2021reframing} traine a supervised filter to select explanations generated by GPT-3 on the SNLI and CommonsenseQA tasks. 


\subsection{Text Classification}
Text classification 
is a task that aims to assign predefined labels (e.g., sentiment polarity, topic, etc) to a given text. 
Earlier work decouple the task into two steps: (1) extract features using neural models such as RNNs~\citep{irsoy2014deep, yang2016hierarchical, wang2018joint, liu2016recurrent, xie2020unsupervised}, CNNs~\citep{Kim2014ConvolutionalNN, zhang2015character, lai2015recurrent, conneau2016very, wei2019eda}, GCN~\citep{yao2019graph}, LLMs~\citep{howard2018universal, sun2019fine, chai2020description, chen2020mixtext, lin2021bertgcn}; and (2) feed extracted features into a classifier~\citep{joulin2016bag} to obtain the final label. 


Recently, in-context learning has achieved success and changes the paradigm in the text classification task. 
\citet{schick2020exploiting} reformulate input examples
into cloze-style phrases and annotate the unlabeled text. 
\citet{han2021ptr} design sub-prompts and applied logic rules to compose sub-prompts into final prompts. 
\citet{liu2021makes} retrieve semantically-similar examples to a test sample to formulate its corresponding prompt. 
\citet{shi2022nearest} retrieve label-words-similar examples as demonstrations in prompts.

\section{Prompt Construction}
\label{sec:methods}
\subsection{Overview}
We follow the standard prompt-based in-context learning paradigm. 
Given an input sequence $\bm{x_{\textit{input}}}=\{x_1, x_2, ..., x_l\}$,
the task of assigning a 
text-class
label to an input text is transformed to 
 generating a pre-defined textual response ${\bm{y}} \in \mathcal{Y}_\textit{verb}$ 
(e.g., positive, negative, etc) 
 conditioning on the prompt $\bm{x}_\textit{prompt}$ using a language model. 
 
 \subsection{Prompt Construction}
 \label{method:prompt}
The prompt $\bm{x}_\textit{prompt}$, which is constructed based on $\bm{x}$, 
consists of the following three components:

\paragraph{(1) Task description $\bm{x}_\textit{desc}$}  generally describes the task. For different classification tasks, e..g, sentiment classification, topic classification, etc, descriptions are different.
Take the sentiment classification task as an example, the task description is given as follows:

{\it Classify the overall sentiment of the input as positive or negative}

\paragraph{(2) Demonstration} consists of a sequence of annotated examples:
$$\{(\bm{x}^{1}_\textit{demo}, \bm{y}^{1}_\textit{demo}), ... , (\bm{x}^{k}_\textit{demo}, \bm{y}^{k}_\textit{demo})\}$$
where $\bm{x}^{j}_\textit{demo}, 1\leq j\leq k$ 
denotes
the $j$th  input sequence 
and $\bm{y}^{j}_\textit{demo}$ denotes the text which is transformed from the label, e.g., positive or negative for the binary sentiment classification task. 
 Demonstration serves as two purposes: (1) providing the LLM with evidence to consult on for decision making, which will significantly boost performances; (2) provides an output format that  LLM's outputs need to follow, so that the output, which takes the form of natural language, can be further easily transformed to labels. 
It is worth noting that demonstrations are only needed for the few-shot learning setup, but not for the zero-shot learning setup. 

\paragraph{(3)  Input $\bm{x_{\textit{input}}}$}  is the test text sequence to classify.

\noindent The prompt  $\bm{x}_\textit{prompt}$  for a test input 
is constructed by concatenating 
the task description $\bm{x}_\textit{desc}$, a sequence of demonstrations $\{(\bm{x}^{1}_\textit{demo}, \bm{y}^{1}_\textit{demo}), ... , (\bm{x}^{k}_\textit{demo}, \bm{y}^{k}_\textit{demo})\}$, and the test sequence $\bm{x}_\textit{test}$,
which can be given as follows:
$$\{\bm{x}_\textit{desc};\text{\textbackslash n};\text{<demo>}^{1};\text{\textbackslash n};...; \text{<demo>}^{k};\text{\textbackslash n};\bm{x}_\textit{test}\}$$

\subsection{Demonstration Sampling}
\label{method:demo_sample}
The few-shot setup requires demonstrations sampled from the training set. 
Strategies that we explore include:

\paragraph{Random Sampling} 
a straightforward strategy from samplings is to randomly sample $k$ examples $\{(\bm{x}^{1}, \bm{y}^{1}), ... , (\bm{x}^{k}, \bm{y}^{k})\}$ from the training set $\mathcal{D}_\textit{train}$ for a text sequence $\bm{x}_\textit{test}$. 

\paragraph{$k$NN Sampling}
The key disadvantage for random sampling is that there is no guarantee that selected samples are semantically related to the input sequence. 
One straightforward alternative is to 
 sample examples that are similar to the test sequence using $k$NN search~\citep{khandelwal2020nearest}. 
 In this process, the test sequence $\bm{x}_\textit{test}$ is first mapped to a vector $\bm{v}_\textit{test}$ using an encoder  model $f$.
Then using $\bm{v}_\textit{test}$ as the query, we search through the entire training set $\mathcal{D}_\textit{train}$ to retrieve $k$ nearest text sequence to get $k$ nearest data examples $\mathcal{N}=\{\bm{x}_j,\bm{y}_j\}_{j=1}^k$ as demonstrations. 
We use the following encoder models to obtain sentence representations and similarity scores:
\paragraph{SimCSE}~\citep{gao2021simcse} is a contrastive learning model for sentence embeddings. 
    We use \texttt{Sup-SimCSE-RoBERTa-Large} model as an encoder model, which is initizlied with RoBERTa-Large~\citep{liu2019roberta} and fine-tuned on the natural language inference datasets. 
    SimCSE~\citep{gao2021simcse} is a semantic-based model and retrieves semantically similar examples, but not necessarily examples with the same labels.
\paragraph{Finetuned Model} FT for short. The key disadvantage for SimCSE~\citep{gao2021simcse} and other general semantic encoding models ~\citep{reimers2019sentence, seonwoo2022ranking,sun2022sentence} is that it measures the general semantic similarity but is not specifically tailored to the text classification task. To resolve this issue, CARP uses the model fine-tuned on the training dataset as the $k$NN encoder model.
Specifically, we first fine-tune a Roberta model on the training data. Next we use the [CLS] embedding as the sentence level representation for KNN search. 
    Since the fine-tuned model is trained based on task-specific labels, 
    it guarantees that retrieved samples are close to the input sequence with respect to the task. 
Using fine-tuned model provides a channel to connect LLMs with the full training set, in spite of the limited number of tokens allowed in demonstrations. This strategy lets the model take the
advantage of both the LLMs' generalization abilities and all task-specific evidence provided by the training dataset.



\section{Clues Collecting and Reasoning}
To enhance the models' reasoning ability in addressing linguistic phenomenon tailored to text classification,
we propose a progressive reasoning strategy that involves clue collection, reasoning and decision making.
This process also mimics how human decisions:
where we first collect evidence from the input, separating chaff from wheat;
next we piece together local evidence to form a global picture, which leads to final decision making. 
Next we first given an overview of the the clue collecting and reasoning process, and then describe implementation details.

\subsection{Overview}


\paragraph{Collecting Clues}
For a test sequence, clues are local fact evidence such as keywords, phrases, contextual information, semantic meaning, semantic relationships, tones, references, etc. 
The following is an example for clues of an input:

\noindent {\bf Input}: {\it Steers turns in a snappy screenplay that curls at the edges; it's so clever you want to hate it.}\\ 
{\bf Clues}: {\it "snappy", "clever", "want to hate it" are clues for determining the sentiment of the input sentence.}

\paragraph{Reasoning}
For reasoning, the LLM is prompted to go  beyond superficial keywords to mine deeper perspectives, considering language phenomenon such as negation, intensification, irony, etc), and piece together local evidence to form the final decision.
The following example shows the reasoning process to decide the sentiment of the above example based on the evidence collected: 

\noindent\textit{1. The phrase "snappy screenplay" implies that the screenplay is of a high quality and is well-crafted.  \\
2. The phrase "curls at the edges" implies that the screenplay is cleverly written. \\
3. The phrase "so clever you want to hate it" is a paradoxical statement, which suggests that the sentiment is positive despite the use of the word "hate". \\
}

\paragraph{Decision Making} Based on the reasoning process, the model makes the decision for the sentiment of the given input: 

\textit{
Overall, the clues and reasoning process point to a positive sentiment for the input sentence.}

The merits for the incorporation of clue finding and reasonings are as follows:
(1) it prompts the model to progressively think and make decisions:  
clue finding focuses more on superficial features such as keywords, while  reasoning  makes deeper justifications 
based on superficial features. This process better mimics how we humans decide;
(2) clue finding and reasoning serve as a tunnel to let human intervene: in the few-shot setup, where clues and 
reasons need to be prepared in advance for  demonstrations,  
we can modify them as we see fit. This is extremely helpful for trouble shooting in the prompt-construction stage for error corrections;
(3) from an interpretation and uncertainty estimation perspective, clues and reasoning in few-shot setups are human-readable influence functions; 
(4) in contrast to list annotated \texttt{(text, label)} pairs in few-shot setups, 
incorporating clues and reasoning process in prompts aligns closer with the {\it instruction tuning} objective. The discrepancy between LLMs training objectives and in-context learning for downstream tasks has been reduced.

\subsection{Collecting clues and reasoning in zero-shot}
In the zero-shot setup, as no demonstration is allowed, 
no concrete example for clues and reasons can be provided. 
In this way, we only add requests asking the model to output clues and reasons in the prompt.
The prompt is given as follows:
\begin{quote}
{\it This is an overall sentiment classifier for opinion snippets. \newline
First, list CLUES (i.e., keywords, phrases, contextual information, semantic relations, semantic meaning, tones, references) for determining the overall sentiment of the input. \newline
Next, deduce a diagnostic reasoning process from clues and the input to determine the overall sentiment. \newline
Finally, determine the sentiment of input as Positive or Negative considering clues, the reasoning process and the input.\newline \newline
INPUT: <text> \newline
CLUES: }   
\end{quote}

\subsubsection{Clue Collecting and Reasoning in few-shot}
In the few-shot setup , we need to prepare clues and reasonings for all examples in the training set in advance as
all training examples 
have chances to be selected as demonstrations given different test inputs. 
Previous efforts in math problems \cite{wei2022chain, Kojima2022LargeLM, ye2022unreliability, zhang2022automatic} prepare hand-drafted reasoning for a few examples, and always use these example as demonstrations. This strategy does not fit for our situation as 
 it is extremely time-intensive to manually  generate clues and reasonings for all training examples,
To resolve this issue,
we harness LLMs for automatic clue and reasoning generation, 
where we ask LLMs to generate clues and reasoning based on both the input and its corresponding label. 

\begin{table*}[t]
\centering
\small
\scalebox{0.8}{
\begin{tabular}{p{4.5cm}p{5cm}p{7cm}}
\toprule
{\bf INPUT \& GOLD LABEL}  & {\bf CLUES} & {\bf REASONING} \\\midrule
{\bf INPUT}: johnnie to and wai ka fai are sure to find an enthusiastic audience among american action adventure buffs, but the film 's interests may be too narrow to attract crossover viewers \newline \textbf{GOLD LABEL}: Negative& 
- Positive Clues: enthusiastic, action, adventure. \newline - Negative Clues: narrow, crossover & 
The diagnostic reasoning process supporting the sentiment determination of the input is that the phrase "johnnie to and wai ka fai are sure to find an enthusiastic audience among american action adventure buffs" suggests that the film is likely to be well-received by this specific demographic. However, the subsequent phrase "but the film's interests may be too narrow to attract crossover viewers" implies that the film is unlikely to appeal to a broader audience, suggesting a negative sentiment. Therefore, the overall sentiment of the input is negative. \\\midrule
{\bf INPUT}: moore sell globe unit moore resources inc said agreed principle sell globe inc unit cyprus minerals co undisclosed terms sale expected close mid april subject certain conditions including board approval companies globe produces silicon metals products used chemical aluminum iron rubber semiconductor industries reuter \newline 
\textbf{GOLD LABEL}: Acquisitions & 
- Acquisition (agreed, sell, unit, sale, close, board approval) \newline 
- Companies (Moore Resources Inc, Globe Inc, Cyprus Minerals Co) \newline 
- Products (silicon, metals, chemical, aluminum, iron, rubber, semiconductor) \newline 
- Reuter (news agency)
& 
The input mentions an acquisition agreement between Moore Resources Inc and Globe Inc, and the sale is expected to close in mid-April, suggesting an Acquisitions topic. The input also mentions Cyprus Minerals Co, silicon and metals products which are used in chemical, aluminum, iron, rubber, and semiconductor industries, and a Reuter news agency, all of which support the Acquisitions topic. \\
\bottomrule
\end{tabular}
}
\caption{Examples of generated clues and reasoning for demonstrations.}
\label{behavior}
\end{table*}

\paragraph{Clue Generation}
For a given training example 
 \textit{<text>} paired with the
  label word  
\textit{<label-word>} (e.g., positive), 
we ask LLM to generate clues that  indicate the label:

\begin{quote}
{\it List  CLUES (i.e., keywords, phrases, contextual information, semantic meaning, semantic relationships, tones, references)  
that support the sentiment determination of the input (limit to 15 words).\newline  
INPUT: <text> \newline
SENTIMENT: <label-word>}    
\end{quote}

\paragraph{Reasoning Generation}
Based on clues generated clues, the input, and the label, 
we ask LLMs to generate reasoning details\footnote{LLMs often generate long responses, in order to ensemble more demonstrations in prompts, we use \textit{"limit to 50 words"}. After conducting an analysis of the generated responses, we find that LLMs can explain the reason within limited words.}:
\begin{quote}
{\it Based on the input and clues, articulate the diagnostic reasoning process that supports the sentiment determination of the input. \newline  
INPUT: <text> \newline
LABEL: <label-word> \newline 
CLUES: <clues> \newline 
REASONING: 
}
\end{quote}

Given the generated clues and reasonings for all training examples, at test time, when K-nearest examples are selected demonstrations, 
its corresponding clues and reasons are concatenated to the demonstration.
In this way, each demonstration example 
is composed by a \texttt{(text, clues, reasons, golden label word)} pair.
The  prompt is thus given as follows:

\begin{quote}
{\it This is a sentiment classifier for input opinion snippets. \newline
List CLUES (i.e., keywords, phrases, contextual information, semantic meaning, semantic relationships, tones, references) that support the sentiment determination of the input. \newline 
Next, deduce the diagnostic REASONING process from premises (i.e., clues, input) that support the sentiment determination. \newline 
Finally, based on clues, the reasoning and the input, categorize the overall SENTIMENT of input as Positive or Negative.\newline \newline
input: <demo-text-1> \newline
clues: <demo-clues-1> \newline 
reasoning: <demo-reason-1> \newline 
sentiment: <demo-label-word-1> \newline 
input: <demo-text-2> \newline
clues: <demo-clues-2> \newline 
reasoning: <demo-reason-2> \newline 
sentiment: <demo-label-word-2> \newline 
... ... \newline 
input: <demo-text-n> \newline
clues: <demo-clues-n> \newline 
reasoning: <demo-reason-n> \newline 
sentiment: <demo-label-word-n> \newline 
input: <text> 
}
\end{quote}

Examples for prompts with clues and reasons are shown in Figure~\ref{fig:few_shot}.
In this way, for a test example,
by following the format of demonstrations, 
the LLM will first output clues, then reasons, and at last decisions. 

\subsection{Voting}
Unlike conventional discriminative models for text classification, which generate deterministic results during inferences, LLMs for in-context learning are generative models and generate distinct textual responses with diverse sampling strategies in multiple runs. 
We consider the following voting strategies in the paper:
\begin{tightitemize}
    \item {\bf Majority Vote}: the final result is the most frequent prediction among multiple runs.  
    \item {\bf Weighted Probability Vote}: the final result is the one with weighted summed probability from multiple runs.
\end{tightitemize}

\begin{table*}[t]
    \small
    \centering
     \scalebox{1.0}{
      \begin{tabular}{lcccccccc}
      \toprule
       & \multicolumn{1}{c}{\bf SST-2}& \multicolumn{1}{c}{\bf  AGNews}& \multicolumn{1}{c}{\bf  R8}&
       \multicolumn{1}{c}{\bf  R52}& 
        \multicolumn{1}{c}{\bf MR} & 
         \multicolumn{1}{c}{\bf Average}
       \\\midrule
      \multicolumn{9}{c}{\underline{\it \bf Supervised Methods}}\vspace{1pt}\\\midrule
      RoBERTa-Large~\citep{liu2019roberta} & 95.99 & 95.55 & 97.76  & 96.42 & 91.16  & 95.38  \\
      DeBERTa~\citep{He2020DeBERTaDB} & 94.75 & 95.32 &  98.33 & 96.32 & 90.19  & 94.99 \\
      RoBERTa-GCN~\citep{lin2021bertgcn}  & 95.80 & \textbf{95.68*} & 98.2  & 96.1 & 89.7  & 95.10  \\
      XLNet~\citep{yang2019xlnet} & \textbf{96.10*} & 95.55   & - & -  & - & -  \\
      VLAWE~\citep{ionescu2019vector} & - & - & - & - & \textbf{93.3*} & - \\
      GCN-SB~\citep{gcnsb} & - & - & \textbf{98.53*} & \textbf{96.35*} & 87.59 & - \\
      \midrule
      \multicolumn{9}{c}{\underline{\it \bf Zero-shot Setting}}\vspace{1pt}\\\midrule
      Vanilla~\citep{brown2020language} & 91.55  & 90.72 & 90.19 & 89.06 & 88.69 & 90.04 \\
      CoT~\citep{Kojima2022LargeLM} & 92.11  & 91.25 & 90.48 & 91.24 & 89.37 & 90.89  \\
      \textbf{CARP} & 93.01  & 92.60 & 91.75 & 91.80 & 89.94 & 91.82  \\\midrule
      \multicolumn{9}{c}{\underline{\it \bf Few-shot Setting ($k$=16)}}\vspace{1pt}\\\midrule
      {\textit{Random Sampler}} &   {\bf }  & {\bf } & {\bf}  & {\bf } & {\bf } & {\bf  } \\\midrule
      Vanilla~\citep{brown2020language} & 92.36 & 91.74 & 91.58 & 91.56 & 89.15 & 91.28  \\
      CoT~\citep{Kojima2022LargeLM} &  94.56 & 95.02 & 92.49 & 92.03 & 89.91 & 92.80   \\
     \textbf{CARP} &  96.20 & 95.18 & 97.60 & 96.19 & 90.03 & 95.04   \\\midrule
    {\textit{SimCSE $k$NN-Sampler}} &   {\bf }  & {\bf } & {\bf}  & {\bf } & {\bf } & {\bf  } \\\midrule
    Vanilla~\citep{brown2020language} & 93.90 & 93.50 & 94.36 & 92.40 & 89.59 & 94.05   \\
    CoT~\citep{Kojima2022LargeLM} & 94.21 & 94.28 & 95.07 & 92.98 & 90.27 & 93.69   \\
    \textbf{CARP} &  95.69  & 95.25 & 97.83 & 96.27 & 90.74 & 95.16   \\\midrule
    {\textit{FT $k$NN-Sampler}} &   {\bf }  & {\bf } & {\bf}  & {\bf } & {\bf } & {\bf  } \\\midrule
   Vanilla~\citep{brown2020language} & 94.01 & 94.14 & 95.57 & 95.79 & 90.90 & 94.08   \\
   CoT~\citep{Kojima2022LargeLM} & 95.48 & 94.89 & 95.59 & 95.89 & 90.17 & 94.40   \\
   \textbf{CARP} & 96.80 & 95.99 & 98.29 & 96.82 & 91.90 & 95.97   \\
   \textbf{CARP} (WP Vote) & 97.39 & 96.40 & 98.78 & 96.95 & 92.39 & 96.38  \\
      \bottomrule
    \end{tabular}
    }
    \caption{Accuracy performances of different settings on benchmarks. 
    We report mean and standard deviation results over 5 runs. 
    The GPT-3 denotes \texttt{text-davinci-003}. 
    In few-shot experiments, we sample 16 annotated examples ($k$=16) for every test instance. 
    \textbf{*} indicates previous state-of-the-art results. "MJ Vote" is short for majority vote. "WP Vote" denotes weighted probability vote. }
    \label{overall-performance}
    \end{table*}

\section{Experiments}
\label{method:experiments}

In order to evaluate the effectiveness of the proposed method, we conduct experiments on two setups:
(1) full training setup, where the model has the access to the full training data; and
(2) low-resource setup, where the model can only access partial 
training
dataset. The low-resource setup better mimics real-world situations where training data is  limited. 
For the full training setup, we follow the standard train/dev/test split.  
For  the low-resource setup, we randomly sample $n$ instances per class ($n$ in $\{16, 128, 256, 512, 1024\}$) from the benchmark training set. 
The sampled subset forms a new training set to test different models' abilities in the low-resource situations. 
During experiments, we train models/sample demonstrations with the new training set. 

We conduct experiments on five
widely-used 
 datasets, including SST-2~\citep{socher2013recursive}, R8, R52\footnote{R8 and R52 are original from \url{https://www.cs.umb.edu/~smimarog/textmining/datasets/}}, AGNews~\citep{zhang2015character} and Movie Review (MR)~\citep{pang2005seeing}. 
More details of the benchmarks and low-resource datasets can be found in Appendix~\ref{app:dataset}. 

For zero-shot and few-shot experiments, we use InstructGPT-3~\citep{ouyang2022training} (\texttt{text-davinci-003}, 175B) as the backbone.
Due to the input token limitation, we use $k=16$ for few-shot setups. 
Prompts on the five datasets are shown in Appendix~\ref{app:prompt}. 
Model hyper-parameters can be found in Table~\ref{tab:hyper-params}~\footnote{During experiments, we find that CARP is robust with different hyper-parameters. Experimental results can be found in Appendix~\ref{app:hyper}}.

We use \textbf{Vanilla} to denote the conventional ICL approach where LLMs are directly prompted to generate labels.
We
use \textbf{CoT}~\citep{Kojima2022LargeLM} to denote the baseline that mimics the chain-of-thought strategy 
and use CARP to denote the proposed method.

\subsection{Models for Comparison}
\paragraph{Supervised models} trained on the trained set naturally constitute baselines to compare with.
We use the following  models as baselines, and more details of hyper-parameters are shown in Appendix~\ref{app:model_hyper}:  
\begin{tightitemize}
    \item {\bf RoBERTa-Large}:We fine-tune RoBERTa-Large~\citep{liu2019roberta} on the training set.
    \item {\bf RoBERTa-GCN}:\citet{lin2021bertgcn} constructs heterogeneous graph networks on top of the RoBERTa-Large~\citep{liu2019roberta} model. 
    \item {\bf DeBERTa}:\citet{He2020DeBERTaDB} improve RoBERTa by using disentangled attention mechanism and an enhanced mask decoder. 
    \item {\bf XLNet}:\citet{yang2019xlnet} propose a generalized autoregressive pretraining method that enables learning bidirectional contexts.
    \item {\bf GCN-SB}:\citet{gcnsb} propose a simplified boosting algorithm, which makes CNN learn the samples misclassified by GCN again. 
    \item {\bf VLAWE}:\citet{ionescu2019vector} obtain document embeddings based on aggregating the differences between each codeword vector and each word vector (from the document) associated to the respective codeword.
\end{tightitemize}

\paragraph{Few-shot Setup}
For demonstration sample strategies in the few-shot setup, we  consider the following strategies for comparison:
 (more details can be found in Section~\ref{method:demo_sample}): 
\begin{tightitemize}
    \item {\bf Random Sampler}: randomly samples $k$ examples. 
    \item {\bf SimCSE $k$NN-Sampler}: samples $k$ nearest examples based on SimCSE~\citep{gao2021simcse} representations\footnote{Specifically, we use \texttt{Sup-SimCSE-RoBERTa-Large} as the text encoder.}.
    \item {\bf FT $k$NN-Sampler}: sample $k$ nearest examples using {\bf F}ine-{\bf T}uned RoBERTa-Large representations. 
\end{tightitemize}

\begin{table}[t]
    \centering
     \scalebox{0.9}{
    \begin{tabular}{lc} \toprule
    {\bf Parameter } &  {\bf Value}   \\\midrule
    Engine Name & text-davinci-003 \\
    Max Tokens & 200 \\ 
    Temperature & 0.7  \\
    Top P & 1 \\
    Frequency Penalty & 0.0 \\
    Presence Penalty & 0.0 \\
    Best Of & 1 \\\bottomrule
    \end{tabular}
    }
    \caption{OpenAI API Hyper-parameters.}
\label{tab:hyper-params}
\end{table}

\subsection{Results on the full training set}
\label{sec:result_full}
Experimental results are shown in Table~\ref{overall-performance}.
As can be seen, performances of few-shot setups consistently outperform 
zero-shot setups. 
In terms of sampling strategies in the few-shot setups, 
we observe that simcse KNN-sampler outperform random sampler, illustrating the importance of adding  demonstrations that are relevant to the test input 
in the few-shot setup.
We also observe that 
 FT KNN-sampler consistently outperforms simcse KNN-sampler.
 This shows that, the fine-tuned model, which takes the advantage of the full training set, serves as a better retriever for task-specific demonstration retrieval than 
 the general-purposed simcse retriever. 

For different reasoning strategies, we first observe that the CoT strategy outperforms 
the vanilla strategy, which straightforwardly asks LLMs to generate results without further reasoning steps. 
CARP consistently outperforms CoT across all benchmarks, i.e., +1.48, +0.97, +2.76, + 3.29, +0.47 respectively on SST-2, AGNews, R8, R52 and MR datasets.
This demonstrates the 
necessity of building models with  complex linguistic phenomena involved in text classification,
and the effectiveness of CARP in doing this job.  

Compared with supervised learning baselines, 
we find that the vanilla model using LLM underperforms supervised baselines, while few-shot CoT is able to obtain 
slightly worse or comparable results agains supervised baselines.
Notably, single CARP outperforms fine-tuned RoBBERTa on all benchmarks.
Using WP voting strategies,
CARP yields new SOTA performances on four
out of the 5 datasets,  97.39 on SST-2 (+1.24),  96.40 (+0.72) on AGNews, 98.78 (+0.25) on R8 and 96.95 (+0.6) on R52,
and a performance comparable to SOTA on MR (92.39 v.s. 93.3). 


\subsection{Results on low-resource settings}
To estimate low-resource circumstances, we sample $n=\{16, 128, 256, 512, 1024\}$ instances for each class as low-resource setups. 
Experimental results are shown in Table~\ref{table:exp_low_resource}. 
As can be seen, when the training set size is extremely small (i.e., 16 or 128 sentences), and the performance of the supervised model is far below CARP. Even with only 16 examples to train on, the accuracy of CARP of SST-2  already around 90\%, whereas supervised models' performance is similar to random guess.
This demonstrates the strong generalization ability of CARP in the low-resource setup.
As we anticipated, the $k$NN search efficiency improved at a faster rate as the amount of the training data increases; 
Enlarging the training dataset increases the chances that the chosen examples will correspond to the input, resulting in improved results.
Specifically, 
using 16 examples per class, CARP achieves comparable performances to supervised models with 1,024 examples per class; 
using 512 instance per class annotation data, 
CARP achieves comparable performances to supervised models trained on the full set.
\begin{table*}[!t]
  \small
  \centering
  \scalebox{1.1}{
  \begin{tabular}{l|lccccc}\toprule
  \multicolumn{1}{l}{\bf Dataset} & 
  \multicolumn{1}{l}{\bf Model} & 
  \multicolumn{1}{c}{\bf $n$=16} & 
  \multicolumn{1}{c}{\bf $n$=128} & 
  \multicolumn{1}{c}{\bf $n$=256} & 
  \multicolumn{1}{c}{\bf $n$=512} & 
  \multicolumn{1}{c}{\bf $n$=1024} 
  \\\midrule 
  & FT RoBERTa & 51.52 & 52.31 & 53.89 & 70.49 & 90.30 \\
 \multirow{2}{*}{\bf SST-2} & \multirow{1}{*}{GPT-3 Vanilla} & 90.15 & 90.36 & 91.70 & 93.86 & 94.68  \\
  & \multirow{1}{*}{GPT-3 Zero-shot-CoT}& 89.66 & 90.19 & 90.80 & 94.42 & 94.89  \\ 
  & \multirow{1}{*}{GPT-3 CRAP} & 90.48 & 91.07 & 91.77 & 94.03 & 95.20 \\\hline
  \multirow{4}{*}{\bf AGNews} & FT RoBERTa & 21.87 & 38.19 & 40.08 & 50.18 & 78.09   \\
 & \multirow{1}{*}{GPT-3 Vanilla} & 89.47 & 89.63 & 90.54 & 93.02 & 94.79  \\
  & \multirow{1}{*}{GPT-3 Zero-shot-CoT}& 89.66 & 90.16 & 91.70 & 94.86 & 95.28  \\ 
  & \multirow{1}{*}{GPT-3 CRAP} & 90.16 & 90.94 & 91.07 & 94.08 & 95.48 \\\hline
  \multirow{4}{*}{\bf R8} & FT RoBERTa & 11.29 & 48.19 & 60.18 & 70.70 & 88.68  \\
& \multirow{1}{*}{GPT-3 Vanilla} & 89.15 & 90.27 & 91.70 & 94.00 & 94.91  \\
  & \multirow{1}{*}{GPT-3 Zero-shot-CoT}& 90.49 & 90.88 & 91.81 & 95.42 & 95.75  \\ 
  & \multirow{1}{*}{GPT-3 CRAP} & 90.23 & 91.03 & 91.77 & 95.56 & 96.67 \\\hline
  \multirow{4}{*}{\bf R52} & FT RoBERTa & 38.29 & 39.10 & 59.18 & 67.19 & 81.53  \\
 & \multirow{1}{*}{GPT-3 Vanilla} & 89.15 & 90.04 & 90.29 & 91.88 & 92.06  \\
  & \multirow{1}{*}{GPT-3 Zero-shot-CoT}& 89.46 & 90.02 & 90.73 & 93.20 & 94.12  \\ 
  & \multirow{1}{*}{GPT-3 CRAP} & 90.82 & 91.00 & 95.85 & 94.36 & 96.27 \\\hline
 \multirow{4}{*}{\bf MR} & FT RoBERTa & 51.20 & 52.11 & 53.58 & 68.29 & 88.37  \\
  & \multirow{1}{*}{GPT-3 Vanilla} & 86.04 & 88.68 & 88.99 & 89.80 & 90.18  \\
  & \multirow{1}{*}{GPT-3 Zero-shot-CoT}& 86.26 & 89.00 & 90.01 & 90.16 & 90.89  \\ 
  & \multirow{1}{*}{GPT-3 CRAP} & 86.54 & 87.19 & 89.63 & 90.01 & 91.20 \\
  \bottomrule
  \end{tabular}
  }
  \caption{Experimental results on low-resource ($n$ example per class) settings. 
  We compare fine-tuned RoBERTa-Large with $16$-shots GPT-3 setting. 
  For GPT-3, we use SimCSE~\citep{gao2021simcse} to retrieve 16 annotated examples from the low-resource train set. 
  "cls" represents GPT-3 makes decisions by generating label words; "reason-cls" denotes that GPT-3 first generates the reasoning process and then makes decisions; "clue-reason-cls" represents that GPT-3 finds clues in the given text, then explain the reasoning process and finally makes decisions. 
  }
  \label{table:exp_low_resource}
\end{table*}

\subsection{Domain Adaptation}
It is unclear whether it is essential to train models on the specific dataset for retrieving demonstrations. 
In this subsection, we conduct an analysis on using  demonstrations from out-of-distribution datasets.

We use  SST-2 and Yelp, and the task is to determine the positive or negative polarity of the given text. 
SST-2 and Yelp are from different domains: 
SST-2 are snippets from Rotten Tomatoes\footnote{\url{https://www.rottentomatoes.com/}}, whereas Yelp\footnote{\url{https://drive.google.com/drive/folders/0Bz8a_Dbh9Qhbfll6bVpmNUtUcFdjYmF2SEpmZUZUcVNiMUw1TWN6RDV3a0JHT3kxLVhVR2M?resourcekey=0-TLwzfR2O-D2aPitmn5o9VQ&usp=share_link}} 
consists of 
 product reviews from the online website. 
\begin{table}[t]
\center
\small
\begin{tabular}{lcc}\toprule 
{\bf } & {\bf FT RoBERTa on} & {\bf FT RoBERTa on} \\
{\bf } & {\bf SST-2 Train} & {\bf Yelp Train}  \\\midrule
SST-2 Test & 95.99  & 88.78  \\
Yelp Test & 92.38 & 96.04 \\
\specialrule{0em}{1pt}{1pt}
\cdashline{1-3}
\specialrule{0em}{1pt}{1pt}
{\bf } & {\bf CARP with} & {\bf CARP with}  \\
{\bf } & {\bf SST-2 demon.} & {\bf Yelp demon.}  \\\midrule
SST-2 Test & 96.80 & 96.29  \\
Yelp Test & 95.94 & 96.32 \\
\bottomrule
\end{tabular}
\caption{Results for Yelp test set when using in-domain/out-of-domain $k$NN sampler and demonstrations source. We use FT $k$NN Sampler to retrieve demonstrations on the corresponding train set. }
\label{ablation:ood}
\end{table}
Experimental results are shown in Table~\ref{ablation:ood}. 
SST-2 train \& SST-2 test means that demonstrations are from the SST-2 dataset and 
test is performed on SST-2 dataset;  
Yelp train \& SST-2 test means demonstrations are from yelp and 
test is performed on SST-2 dataset.
We see a significant decrease (-7.2\%, 95.99\% v.s.88.78\% ) in performance when switching SST-2 train to Yelp-2 train using supervised RoBERTa, which illustrates that supervised models are very sensitive 
to the out-of-distribution data.
On the contrary, 
we only observe a slight decrease in performance (-0.5\%, 96.80\% v.s. 96.29\%) when switching SST-2 train to Yelp-2 train on SST-2 test,
illustration the greater capabilities of CARP on the domain adaptation situations.

This means CARP is very robust when training and test are not from the same domain.
On the contrary, 

\begin{table*}[t]
    \small
    \centering
     \scalebox{0.95}{
      \begin{tabular}{lcccccccc}
      \toprule
       & \multicolumn{1}{c}{\bf SST-2}& \multicolumn{1}{c}{\bf  AGNews}& \multicolumn{1}{c}{\bf  R8}&
       \multicolumn{1}{c}{\bf  R52}& 
        \multicolumn{1}{c}{\bf MR} & 
         \multicolumn{1}{c}{\bf Average}
       \\\midrule
      \multicolumn{9}{c}{\underline{\it \bf Supervised Methods}}\vspace{1pt}\\\midrule
      RoBERTa-Large & 95.99 & 95.55 & 97.76  & 96.42 & 91.16  & 95.38 \\
      RoBERTa-GCN  & 95.80 & 95.68 & 98.2  & 96.1 & 89.7  & 95.10  \\
      \midrule
      \multicolumn{9}{c}{\underline{\it \bf Zero-shot Setting}}\vspace{1pt}\\\midrule
      Vanilla  & 91.55  & 90.72 & 90.19 & 89.06 & 88.69 & 90.04 \\
      Zero-shot-CoT & 92.11  & 91.25 & 90.48 & 91.24 & 89.37 & 90.89  \\
      \textbf{CARP} & 94.41  & 93.18 & 93.29 & 92.69 & 90.03 & 92.72   \\\midrule
      \multicolumn{9}{c}{\underline{\it \bf Few-shot Setting}}\vspace{1pt}\\\midrule
      {\textit{Random Sampler}} &   {\bf }  & {\bf } & {\bf}  & {\bf } & {\bf } & {\bf  } \\\midrule
      Vanilla &  91.36 & 91.48 & 90.60 & 90.68 & 89.15 & 90.65   \\
      Zero-shot-CoT &  92.56 & 92.65 & 92.49 & 92.03 & 89.91 & 91.93  \\
     \textbf{CARP} &  94.41  & 93.18 & 93.29 & 92.69 & 90.03 & 92.72   \\\midrule
    {\textit{SimCSE $k$NN-Sampler}} &   {\bf }  & {\bf } & {\bf}  & {\bf } & {\bf } & {\bf  } \\\midrule
    Vanilla &  93.90 & 93.50 & 94.36 & 92.40 & 89.59 & 92.75   \\
    Zero-shot-CoT &  94.21 & 94.28 & 95.07 & 92.98 & 90.27 & 93.36     \\
    \textbf{CARP} &  95.99  & 95.53 & 95.31 & 93.84 & 90.64 & 94.26   \\\midrule
    {\textit{FT $k$NN-Sampler}} &   {\bf }  & {\bf } & {\bf}  & {\bf } & {\bf } & {\bf  } \\\midrule
   Vanilla & 94.01 & 94.14 & 95.57 & 95.79 & 90.90 & 94.08   \\
   Zero-shot-CoT & 95.48 & 94.89 & 95.59 & 95.89 & 90.17 & 94.40   \\
   \textbf{CARP} & 96.62 & 95.97 & 98.13 & 96.12 & 91.86 & 95.74   \\
      \bottomrule
      \end{tabular}
      }
    \caption{Accuracy performances of different settings on test subsets (results are over 5 runs). 
    GPT-3 denotes \texttt{text-davinci-003}. 
    In few-shot experiments, we sample 16 annotated examples ($k$=16) per prompt. "MJ Vote" is short for majority vote. "WP Vote" denotes weighted probability vote.}
    \label{subset-performance}
    \end{table*}
\section{Ablation Studies}
In this section, we conduct comprehensive ablation studies to get a better knowledge about different elements of CARP. 

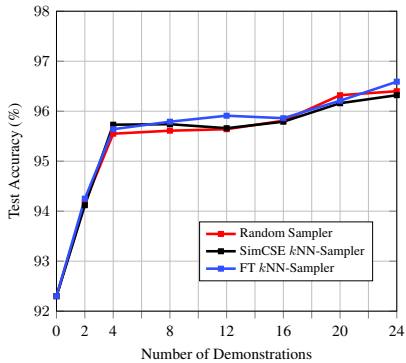
\begin{figure}[!ht]     
\centering
\begin{tikzpicture}[scale=0.65]
\begin{axis}[
    	   width=1.1\columnwidth,
	    height=1\columnwidth,
	    legend cell align=left,
	    legend style={at={(0.93, 0.1)},anchor=south east,font=\small,nodes={scale=0.85, transform shape}},
	    xtick={0,1,2,3,4,5,6,7,8,9,10, 11, 12},
	    xticklabels={0, 2, 4, , 8, , 12, , 16, ,20, , 24},
   		ytick={92, 93, 94, 95, 96, 97, 98},
        ymin=92, ymax=98,
   		xtick pos=left,
   		xtick align=outside,
	    xmin=0,xmax=12,
	    mark options={mark size=1},
		font=\small,
   	 	ymajorgrids=true,
    	xmajorgrids=true,
    	xlabel=Number of Demonstrations,
        ylabel=Test Accuracy (\%),
    	ylabel style={at={(axis description cs: -0.08, 0.5)}}]
    	
\addplot[
    color=red,
    mark=square*,
    line width=1.5pt
    ]
    coordinates {
(0, 92.30)(1, 94.17)(2, 95.55)(4,95.61)(6,95.64)(8,95.81)(10,96.32)(12,96.40)
    };
    \addlegendentry{Random Sampler}

\addplot[
    color=black,
    mark=square*,
    line width=1.5pt
    ]
    coordinates {
(0, 92.30)(1, 94.12)(2, 95.73)(4,95.74)(6,95.66)(8,95.79)(10,96.16)(12,96.32)
    };
    \addlegendentry{SimCSE $k$NN-Sampler}

\addplot[
    color=shannon,
    mark=square*,
    line width=1.5pt
    ]
    coordinates {
(0, 92.30)(1, 94.25)(2, 95.64)(4, 95.79)(6, 95.91)(8, 95.86)(10, 96.21)(12, 96.59)
    };
    \addlegendentry{FT $k$NN-Sampler}

\end{axis}
\end{tikzpicture}
\caption{Performances v.s. the number of demonstrations in few-shot prompts.}
\label{fig:num_demo}
\end{figure}

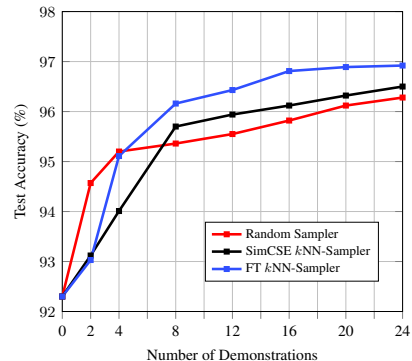
\begin{figure}[!ht]     
\centering
\begin{tikzpicture}[scale=0.65]
\begin{axis}[
    	   width=1.1\columnwidth,
	    height=1\columnwidth,
	    legend cell align=left,
	    legend style={at={(0.93, 0.1)},anchor=south east,font=\small,nodes={scale=0.85, transform shape}},
	    xtick={0,1,2,3,4,5,6,7,8,9,10, 11, 12},
	    xticklabels={0, 2, 4, , 8, , 12, , 16, ,20, , 24},
   		ytick={92, 93, 94, 95, 96, 97, 98},
        ymin=92, ymax=98,
   		xtick pos=left,
   		xtick align=outside,
	    xmin=0,xmax=12,
	    mark options={mark size=1},
		font=\small,
   	 	ymajorgrids=true,
    	xmajorgrids=true,
    	xlabel=Number of Demonstrations,
        ylabel=Test Accuracy (\%),
    	ylabel style={at={(axis description cs: -0.08, 0.5)}}]
    	
\addplot[
    color=red,
    mark=square*,
    line width=1.5pt
    ]
    coordinates {
(0, 92.30)(1, 94.57)(2, 95.20)(4, 95.36)(6, 95.55)(8, 95.82)(10,96.12)(12,96.28)
    };
    \addlegendentry{Random Sampler}

\addplot[
    color=black,
    mark=square*,
    line width=1.5pt
    ]
    coordinates {
(0, 92.30)(1, 93.12)(2, 94.01)(4,95.70)(6,95.94)(8,96.12)(10,96.32)(12,96.50)
    };
    \addlegendentry{SimCSE $k$NN-Sampler}

\addplot[
    color=shannon,
    mark=square*,
    line width=1.5pt
    ]
    coordinates {
(0, 92.30)(1, 93.03)(2, 95.11)(4, 96.16)(6, 96.43)(8, 96.81)(10, 96.89)(12, 96.92)
    };
    \addlegendentry{FT $k$NN-Sampler}

\end{axis}
\end{tikzpicture}
\caption{Performances v.s. the number of demonstrations in few-shot prompts for the CARP strategy, where LLMs are first asked to generate evidence, then to reason and at last to generate final results.}
\label{fig:cr_num_demo}
\end{figure}

\subsection{Impact of the number of demonstrations}
We explore the effect of the number of demonstrations in prompts. We conduct experiments on the SST-2 dataset. 
Results for the vanilla prompting and the CARP schemas using different sampling strategies are shown in Figure~\ref{fig:num_demo} and Figure~\ref{fig:cr_num_demo}, respectively. 
As can be seen, performances  improve as the number of demonstrations increases for both the vanilla and the CARP schemas. 




\begin{table}[t]
\center
\small
\scalebox{0.9}{
\begin{tabular}{lcc}\toprule 
{\bf Prompts} & {\bf SST-2} &  {\bf R8} \\\midrule
CARP & 96.80 & 98.29 \\\midrule
w/o Text & 92.28 & 94.18 \\
w/o Clue & 95.48 & 95.29\\
w/o Reason & 95.72 & 97.82 \\
w/o Label & 96.53 & 98.18 \\
\bottomrule
\end{tabular}
}
\caption{The effect of components on the SST-2 dataset with different strategies.}
\label{ablation:compent_prmompt}
\end{table}

\subsection{The effect of components in demonstrations}
CARP uses \texttt{(text, clues, reasons, golden label word)} pairs as demonstrations. 
In this subsection, we exploit the influence of each component in \texttt{(text, clues, reasons, golden label word)} by removing it from prompts. 
Experimental results are shown in Table~\ref{ablation:compent_prmompt}. 
As shown in Table~\ref{ablation:compent_prmompt}, text in demonstrations has the biggest influence impact of the final results. 
When \texttt{(text, clue, reason)} as demonstrations, the \texttt{label} has effect to the performances. 

\subsection{The effect of different types of label words} 
Label words denote words generated by LLMs 
that indicate the label of the input.  
In this subsection, we explore the impact of using different kinds of label words: 
\begin{tightitemize}
    \item {\bf Position index}: number of index. i.e., one, two, three and etc to denote the label.
    \item {\bf Annotation words}: words used to refer to the category in the annotation file. e.g., positive, negative.~\footnote{GPT-3 generates the same label words for binary sentiment classification task.}
    \item {\bf Synonyms words}: synonyms words e.g., great, terrible. 
    \item {\bf Flipped words}: words that are contrary to original target meanings. 
    e.g., "positive" to denote the negative polarity, "negative" to denote the positive polarity. 
    \item {\bf Random words}: randomly choose words in the vocabulary. e.g., order, number. 
    \item {\bf Special tokens}: tokens that do not have semantic meaning. They are independent of the input and added for a certain purpose. e.g., <cls>, <mask>. 
\end{tightitemize}

\begin{table}[t]
\center
\small
\scalebox{0.9}{
\begin{tabular}{llll}\toprule 
{\bf Strategy} & {\bf Label Words(+,-)} & {\bf CARP} \\\midrule
Position Index & One, Two  & 95.66\\
Annotation Words & Positive, Negative  & {\bf 96.86}  \\
Synonyms Words & Great, Terrible  & 96.27 \\
Flipped Words & Negative, Positive & 64.63\\
Random Words & Cf, Ng  & 95.06\\
Special Tokens & <POS>, <NEG>  & 96.65\\
\bottomrule
\end{tabular}
}
\caption{Label words and results on the SST-2 dataset with different strategies.
"+" represents positive polarity; "-" denotes negative polarity.}
\label{ablation:label_words}
\end{table} 

Results  are shown in Table~\ref{ablation:label_words}.
As can be seen, few-shot ICL with annotation words as label words achieves the best performances.
It is also worth noting that we observe a significant performance  decrease when 
flipped words are used as label words in demonstrations. 

\subsection{The influence of clues}
As mentioned in Section~\ref{sec:methods}, clues are keywords, phrases, contextual information, semantic meaning, semantic relationships, tones, references that support making decisions. 
We remove different types of words in clues and evaluate its influence on  SST-2 and R8 datasets. 
Editing prompts achieve this goal. 
The original prompt for clue collecting is {\it
List CLUES (i.e., keywords, phrases, contextual information, semantic meaning, semantic relationships, tones, references) that support the sentiment determination of the input.} 
If we want to remove {\it keywords \& phrases}, we just remove them from the prompt. 

\begin{tightitemize}
\item {\bf w/o keywords \& phrases}: keywords and phrases are surface evidence for making decisions such as \textit{"like"}, \textit{"hate"}. 
\item {\bf w/o contextual information \& semantic meaning}: contextual information and semantic meaning are meaning in sentences/paragraphs such as \textit{The author express his happiness}. 
\item {\bf w/o semantic relationships}: semantic relationships refer to relations between subjects such as \textit{"emotional danger" suggests a romantic and thrilling relationship between Idemoto and Kim that creates a positive sentiment.}. 
\item {\bf w/o tones}: tones are the general mood of the text such as \textit{The sentence is expressed in an objective tone}. 
\item {\bf w/o references}: references are mentions of commonsense facts or books such as \textit{The reference to the popular, comedic character "Ferris Bueller" implies that the kid is seen in a positive light.}.
\end{tightitemize}

\begin{table}[t]
\center
\small
\scalebox{0.9}{
\begin{tabular}{lcc}\toprule 
{\bf Prompts} & {\bf SST-2} &  {\bf R8} \\\midrule
Clues & 96.80 & 98.29 \\\midrule
w/o keyword\&phrase & 96.21 & 96.91 \\
w/o contextual info. & 96.23 & 97.10 \\
w/o semantic relations & 96.30 & 97.38 \\
w/o tones & 96.40 & 97.35\\
w/o reference & 96.50 & 97.19 \\
\bottomrule
\end{tabular}
}
\caption{Label words and results on the SST-2 dataset with different strategies.}
\label{ablation:clues}
\end{table} 

Experimental results are shown in Table~\ref{ablation:clues}. 
For R8 and SST-2 datasets, keywords play the key role for GPT predictions.


\subsection{The effect of demonstration order}
During experiments, we find that the ranking order of demonstration affect  final results. 
In this subsection, we further investigate the influence of orders of demonstrations. As mentioned in Section~\ref{method:demo_sample}, we retrieved $k$ data instances $\mathcal{N}=\{\bm{x}_j,\bm{y}_j\}_{j=1}^k$ according to the cosine similarity with the test sequence. 
Orders the demonstrations in the prompt we investigate include: 
\begin{tightitemize}
    \item {\bf Random}: randomly shuffle retrieved demonstrations. 
    \item {\bf Low-to-High}: 
    demonstrations with lower similarity scores come first. Therefore
    demonstrations with higher similarity scores are placed closer to the test sequence, which is placed at the end of the prompt. 
    \item {\bf High-to-Low}: demonstrations with lower similarity scores are placed closer to the test sequence.
\end{tightitemize}
\begin{table}[t]
\center
\small
\begin{tabular}{lll}\toprule 
{\bf Ranking} & {\bf SimCSE} & {\bf FT} \\\midrule
\multicolumn{3}{c}{{\it \bf CARP}}\vspace{1pt}\\
\specialrule{0em}{1pt}{1pt}
\cdashline{1-3}
\specialrule{0em}{1pt}{1pt}
Random &  95.39	& 95.99 \\
High-to-Low & 95.22	& 96.71 \\
Low-to-High & 96.39 & 96.80 \\\bottomrule
\end{tabular}
\caption{Accuracy scores on SST-2 when assembling demonstrations with different ranking strategies.}
\label{ablation:rank_demo}
\end{table}

As shown in Table~\ref{ablation:rank_demo}, 
 performance is sensitive the ordering of the demonstrations.
The low-to-high ordering  achieves the best performance compared to random and high-to-low ordering.

\subsection{Quality of the reasoning process} 
\label{study:quality}
In this paper, we use LLMs to generate rationable explanations instead of human editing. 
Therefore, the quality of generated reasoning process affects the final results. 
In this subsection, we sample 500 training \textit{(text, clues, reason, label)} pairs and evaluate the 
 generated reasoning process from the following perspectives:  

\textbf{(1) Reliability:}~Inspired by the emergent generalization ability of LLMs, we use zero-shot GPT-3 (175B) as the self-critique model to evaluate the quality of generated reasoning processes. To be specific, we ask the GPT-3 to return yes/no if the generated reasoning process supports making decisions for the input text.
If the GPT-3 returns "yes", it denotes that the reasoning process is reliable for 
making decisions. If the GPT-3 returns "no", it represents that the reasoning process is not reliable.

The prompt for SST-2 is shown as follows: 

\noindent \textit{Is the following REASONING process supporting determinate sentiment label to INPUT? Please answer Yes or No. \\
INPUT: <text> \\
REASONING: <reasoning-process>
}

where \textit{<text>} is the text sequence for the data and \textit{<reasoning-process>} is generated reasoning process.

\begin{table}[t]
\center
\small
\scalebox{0.7}{
\begin{tabular}{lccc}\toprule 
{\bf } & {\bf Reliability(\%)~$\uparrow$} & {\bf Fluency(ppl)~$\downarrow$} & {\bf Logic Faithful(\%)~$\uparrow$}\\\midrule
SST-2 &  96.18  & 3.89 &  95.20  \\
R8 & 95.34 & 3.29 & 94.55 \\
\bottomrule
\end{tabular}
}
\caption{Results for evaluating the quality of generated reasoning explanation. We sample 500 (text, reason) instances for SST-2 and R8.}
\label{ablation:quality_reason}
\end{table}

\textbf{(2) Fluency:}~use LLMs to generate reasoning explanations is a reference-free text generation task. We use perplexity to evaluate the generated text.

\textbf{(3) Logic Faithful:}~previous work often use models, which are trained on natural language inference datasets, to determine whether the given “hypothesis” logically follows from the “premise”.
However, lacking annotation datasets, NLI-trained models can not generalize across multiple domains (e.g., opinion, reviews, news).
Since then, we use $16$-shot ICL with GPT-3 to evaluate whether the generated rationable explanations can be entailed from the input text. 
If the InstructGPT responds with "entailment", it denotes that the generated reasoning process is logic faithful with the text. Otherwise, it represents the reasoning process is not faithful to the text. 
We sample training instances from the SNLI dataset~\citep{snli:emnlp2015} as demonstrations. 
And prompts are shown as follows: 

\noindent \textit{
Given the premise and hypothesis, please justify whether the HYPOTHESIS can be entailed from the PREMISE. Please return yes or no. \newline  
PREMISE: <text> \newline
HYPOTHESIS: <reasoning-process>
}

Evaluation results are shown in Table~\ref{ablation:quality_reason}. 
As can be seen, the reliability percentages for SST-2 and R5 are higher than 95\%.
This indicates that it is feasible to use the model-generated reasoning process as part of the prompts to augment ICL performances.
The perplexity of generated reasoning text is smaller than 4, which denotes that the generated reasoning text is fluent. 
And scores of logic faithful are larger than 93\%, which is in line with our expectation that LLMs can generate reasonable explanations.

\section{Conclusion} 
In this paper, we introduce \textbf{C}lue \textbf{A}nd \textbf{R}easoning \textbf{P}rompting (CARP) for text classification task.  
CARP yields new SOTA performances on 4 out of 5 widely-used
text-classification benchmarks. 
More importantly, we find that CARP delivers impressive abilities on 
low-resource and domain-adaption setups.
In the future, we would like to explore CARP on more natural language understanding tasks.  


\bibliography{custom}
\bibliographystyle{acl_natbib}

\appendix

\section{Dataset}
SST-2~\citep{socher2013recursive}, R8, R52\footnote{R8 and R52 are from \url{https://www.cs.umb.edu/~smimarog/textmining/datasets/}}, AGNews~\citep{zhang2015character} and MR (Movie Review)~\citep{pang2005seeing}. 

\begin{tightitemize}
    \item {\bf SST-2}: The original data in SST-2 are sampled from snippets of Rotten Tomatoes HTML files. 
We use the same train/dev/test splits with \citet{socher2013recursive}. 
    \item {\bf R8 and R52}: R8 and R5211 are two subsections of the Reuters collection, containing 8 and 52 classifications, respectively. The R8 dataset is composed of 5,485 documents for training and 2,189 documents for testing. The R52 dataset is composed of 6,532 training and 2,568 test documents.
    \item {\bf AGNews}: The AG News consists of news articles from the AG’s corpus. The dataset contains 30,000 training and 1,900 testing examples for each class.
    \item {\bf MR (Movie Review)}: The MR contains reviews of films for determining whether a sentiment is either positive or negative.  The corpus has 10,662 reviews. We follow \citep{tang2015pte} and use the same train/test split.
\end{tightitemize}

\begin{table}[t]
\center
\small
\scalebox{0.8}{
\begin{tabular}{l|lcc|ccc}\hline
{\bf Dataset}  & {\bf Task}& {\bf \# Label} & {\bf Source} & {\bf \# Train} & {\bf \# Dev} & {\bf \# Test} \\\hline
SST-2 & sentiment & 2 & review & 6,920 & 872 & 1,821   \\
AGNews & topic & 4 & news & 96,000 & 24,000 & 7,600   \\
R8 & topic & 8 & news & 4,941 & 544 & 2,189  \\
R52 & topic & 52 & news & 5,905 & 627 & 2,568   \\
MR & sentiment & 2 & reviews & 6,398 & 710 & 3,554   \\\hline
\end{tabular}
}
\caption{Benchmark Dataset}
\label{appendix:dataset}
\end{table}
\begin{table}[t]
\center
\small
\scalebox{0.8}{
\begin{tabular}{l|lcc|ccc}\hline
{\bf Dataset}  & {\bf Task}& {\bf \# Label} & {\bf Source} & {\bf \# Train} & {\bf \# Dev} & {\bf \# Subtest} \\\hline
SST-2 & sentiment & 2 & review & 6,920 & 872 & 728  \\
AGNews & topic & 4 & news & 96,000 & 24,000 & 760  \\
R8 & topic & 8 & news & 4,941 & 544 & 875 \\
R52 & topic & 52 & news & 5,905 & 627 & 1,027  \\
MR & sentiment & 2 & reviews & 6,398 & 710 & 888   \\\hline
\end{tabular}
}
\caption{Dataset Subsets}
\label{appendix:dataset}
\end{table}





\section{Hyper-parameters}

\subsection{Fine-tuning Hyper-parameters}
\label{app:model_hyper}
We fine-tune RoBERTa and RoBERT-GCN on 4 NVIDIA 3090 GPUs with FP16.
Model hyper-parameters are tuned on the validation set, where 
learning rate $\{$2e-5, 3e-5, 4e-5$\}$, batch size $\{16, 32, 32\}$, 
a dropout rate of 0.3, a weight decay of 0.01, 
a warmup proportion of 0.01.

\subsection{The influence of hyper-parameters}
\label{app:hyper}
We investigate the effect of model hyper-parameters including temperature, frequency penalty. 
We conduct experiments with Instruct-GPT3 on the SST-2 dataset. 

\paragraph{Temperature}
The temperature $\tau$ controls the generated text variety when another hyper-parameter $top_p$=1. More higher $\tau$, more variety is introduced. 
When $\tau$ is close to 0, the model generates the same result with the greedy decoding method. 
To exploit the effect of temperature $\tau$, we set $\tau$ from 0 to 1.0.
Experimental results are shown in Table~\ref{ablation:temperature}. 
We tokenize the response text with GPT-Tokenizer\footnote{\url{https://platform.openai.com/tokenizer}} and then count the number of tokens. 

\begin{center}
\small
\begin{tabular}{cccl}\hline
{\bf $\tau$}&{\bf SST-2 Accuracy}\\\hline
$\tau=0.0$& 96.39 \\
$\tau=0.2$& 96.48 \\
$\tau=0.4$& 96.40 \\
$\tau=0.6$& 96.59 \\
$\tau=0.8$& 96.68 \\
$\tau=1.0$& 96.70 \\\hline
\end{tabular}
\label{ablation:temperature}
\end{center}






\begin{table*}
\label{tab:methods}
\centering
\small
\vskip 0.1in
\scalebox{0.95}{
\begin{tabular}{ll}
    \toprule
    \multicolumn{2}{c}{\bf SST-2 : positive/negative sentiment analysis} \\
    \toprule
    {\bf \textit{Label Word Map}} & \{0: Negative, 1: Positive\} \\\midrule
    {\bf \textit{Zero-Shot}} & \\
    \specialrule{0em}{1pt}{1pt}
    \cdashline{1-2}
    \specialrule{0em}{1pt}{1pt}
    Classify Prompt: & Please classify the overall SENTIMENT polarity of the INPUT sentence as Positive or Negative. \\ 
    & INPUT: <sent> \\
    & SENTIMENT:  \\
    \specialrule{0em}{1pt}{1pt}
    \cdashline{1-2}
    \specialrule{0em}{1pt}{1pt}
    Reason-Classify Prompts: & Please classify the overall SENTIMENT polarity of the INPUT sentence as Positive or Negative. \\
    & INPUT: <sent> \\
    & \text{} \\
    \specialrule{0em}{1pt}{1pt}
    \cdashline{1-2}
    \specialrule{0em}{1pt}{1pt}
    Findclue-Reason-Classify & {\bf Step 1:} \\
    & Please classify the overall SENTIMENT polarity of the INPUT sentence as Positive or Negative.\\ 
    & INPUT: <sent> \\
    & \text{} \\
    & {\bf Step 2:}\\ 
    & Please classify the overall SENTIMENT polarity of the INPUT sentence as Positive or Negative. \\ 
    & INPUT: <sent> \\
    & CLUES: <step-1-response>\\
    & \text{} \\
    \midrule
    {\bf \textit{Few-Shot}} & \\
    \specialrule{0em}{1pt}{1pt}
    \cdashline{1-2}
    \specialrule{0em}{1pt}{1pt}
    Classify Prompt: & Please classify the overall SENTIMENT polarity of the INPUT sentence as Positive or Negative. \\ 
    & INPUT: <demo-sent> \\
    & SENTIMENT: <demo-label-word> \\
    & \text{} \\
    & INPUT: <demo-sent> \\
    & SENTIMENT: <demo-label-word> \\
    & \text{} \\
    & INPUT: <sent> \\
    & SENTIMENT:  \\
    \specialrule{0em}{1pt}{1pt}
    \cdashline{1-2}
    \specialrule{0em}{1pt}{1pt}
    Reason-Classify Prompts: &  {\bf Step 1:}\\
    & Classify the sentiment of the input sentence as positive or negative. \\
    & INPUT: <demo-sent> \\
    & \text{} \\
    & {\bf Step 2:} \\ 
    & Classify the sentiment of the input sentence as positive or negative.  \\
    & \text{} \\
    & INPUT: <demo-sent> \\
    & REASONING: <step-1-generated>\\ 
    & SENTIMENT: <demo-label-word>\\
    & \text{} \\
    & INPUT: <demo-sent> \\
    & REASONING: <step-1-generated>\\ 
    & SENTIMENT: <demo-label-word>\\
    & \text{} \\
    & INPUT: <test-sent> \\
    \specialrule{0em}{1pt}{1pt}
    \cdashline{1-2}
    \specialrule{0em}{1pt}{1pt}
    Findclue-Reason-Classify Prompts: &  {\bf Step 1:}\\
    & Classify the sentiment of the input sentence as positive or negative. \\
    & INPUT: <demo-sent> \\
    & \text{} \\
    & {\bf Step 2:} \\ 
    & Classify the sentiment of the input sentence as positive or negative.  \\
    & \text{} \\
    & INPUT: <demo-sent> \\
    & REASONING: <step-1-generated>\\ 
    & SENTIMENT: <demo-label-word>\\
    & \text{} \\
    & INPUT: <demo-sent> \\
    & REASONING: <step-1-generated>\\ 
    & SENTIMENT: <demo-label-word>\\
    & \text{} \\
    & INPUT: <test-sent> \\
    \bottomrule
\end{tabular}
}
\caption{Examples of prompts for setups in Section~\ref{sec:methods}.}
\label{example:prompt_sst2}
\end{table*}

\begin{table*}
\label{tab:methods}
\centering
\small
\vskip 0.1in
\scalebox{0.95}{
\begin{tabular}{ll}
    \toprule
    \multicolumn{2}{c}{\bf R8 : topic classification} \\
    \toprule
    {\bf \textit{Label Word Map}} & \{0: Money/Foreign Exchange,
      1: Acquisitions,
      2: Trade,
      3: Interest Rates, \\
      & 4: Shipping,
      5: Earnings and Earnings Forecasts,
      6: Grain,
      7: Crude Oil\} \\\midrule
    {\bf \textit{Zero-Shot}} & \\
    \specialrule{0em}{1pt}{1pt}
    \cdashline{1-2}
    \specialrule{0em}{1pt}{1pt}
    Classify Prompt: & Please classify the overall SENTIMENT polarity of the INPUT sentence as Positive or Negative. \\ 
    & INPUT: <sent> \\
    & SENTIMENT:  \\
    \specialrule{0em}{1pt}{1pt}
    \cdashline{1-2}
    \specialrule{0em}{1pt}{1pt}
    Reason-Classify Prompts: & Please classify the overall SENTIMENT polarity of the INPUT sentence as Positive or Negative. \\
    & INPUT: <sent> \\
    & \text{} \\
    \specialrule{0em}{1pt}{1pt}
    \cdashline{1-2}
    \specialrule{0em}{1pt}{1pt}
    Findclue-Reason-Classify & {\bf Step 1:} \\
    & Please classify the overall SENTIMENT polarity of the INPUT sentence as Positive or Negative.\\ 
    & INPUT: <sent> \\
    & \text{} \\
    & {\bf Step 2:}\\ 
    & Please classify the overall SENTIMENT polarity of the INPUT sentence as Positive or Negative. \\ 
    & INPUT: <sent> \\
    & CLUES: <step-1-response>\\
    & \text{} \\
    \midrule
    {\bf \textit{Few-Shot}} & \\
    \specialrule{0em}{1pt}{1pt}
    \cdashline{1-2}
    \specialrule{0em}{1pt}{1pt}
    Classify Prompt: & Please classify the overall SENTIMENT polarity of the INPUT sentence as Positive or Negative. \\ 
    & INPUT: <demo-sent> \\
    & SENTIMENT: <demo-label-word> \\
    & \text{} \\
    & INPUT: <demo-sent> \\
    & SENTIMENT: <demo-label-word> \\
    & \text{} \\
    & INPUT: <sent> \\
    & SENTIMENT:  \\
    \specialrule{0em}{1pt}{1pt}
    \cdashline{1-2}
    \specialrule{0em}{1pt}{1pt}
    Reason-Classify Prompts: &  {\bf Step 1:}\\
    & Classify the sentiment of the input sentence as positive or negative. \\
    & INPUT: <demo-sent> \\
    & \text{} \\
    & {\bf Step 2:} \\ 
    & Classify the sentiment of the input sentence as positive or negative.  \\
    & \text{} \\
    & INPUT: <demo-sent> \\
    & REASONING: <step-1-generated>\\ 
    & SENTIMENT: <demo-label-word>\\
    & \text{} \\
    & INPUT: <demo-sent> \\
    & REASONING: <step-1-generated>\\ 
    & SENTIMENT: <demo-label-word>\\
    & \text{} \\
    & INPUT: <test-sent> \\
    \specialrule{0em}{1pt}{1pt}
    \cdashline{1-2}
    \specialrule{0em}{1pt}{1pt}
    Findclue-Reason-Classify Prompts: &  {\bf Step 1:}\\
    & Classify the sentiment of the input sentence as positive or negative. \\
    & INPUT: <demo-sent> \\
    & \text{} \\
    & {\bf Step 2:} \\ 
    & Classify the sentiment of the input sentence as positive or negative.  \\
    & \text{} \\
    & INPUT: <demo-sent> \\
    & REASONING: <step-1-generated>\\ 
    & SENTIMENT: <demo-label-word>\\
    & \text{} \\
    & INPUT: <demo-sent> \\
    & REASONING: <step-1-generated>\\ 
    & SENTIMENT: <demo-label-word>\\
    & \text{} \\
    & INPUT: <test-sent> \\
    \bottomrule
\end{tabular}
}
\caption{Examples of prompts for setups in Section~\ref{sec:methods}.}
\label{example:prompt_r8}
\end{table*}

\begin{table*}
\label{tab:methods}
\centering
\small
\vskip 0.1in
\scalebox{0.95}{
\begin{tabular}{ll}
    \toprule
    \multicolumn{2}{c}{\bf MR : topic classification} \\
    \toprule
    {\bf \textit{Label Word Map}} & \{0: Negative, 1: Positive\} \\\midrule
    {\bf \textit{Zero-Shot}} & \\
    \specialrule{0em}{1pt}{1pt}
    \cdashline{1-2}
    \specialrule{0em}{1pt}{1pt}
    Classify Prompt: & Please classify the overall SENTIMENT polarity of the INPUT sentence as Positive or Negative. \\ 
    & INPUT: <sent> \\
    & SENTIMENT:  \\
    \specialrule{0em}{1pt}{1pt}
    \cdashline{1-2}
    \specialrule{0em}{1pt}{1pt}
    Reason-Classify Prompts: & Please classify the overall SENTIMENT polarity of the INPUT sentence as Positive or Negative. \\
    & INPUT: <sent> \\
    & \text{} \\
    \specialrule{0em}{1pt}{1pt}
    \cdashline{1-2}
    \specialrule{0em}{1pt}{1pt}
    Findclue-Reason-Classify & {\bf Step 1:} \\
    & Please classify the overall SENTIMENT polarity of the INPUT sentence as Positive or Negative.\\ 
    & INPUT: <sent> \\
    & \text{} \\
    & {\bf Step 2:}\\ 
    & Please classify the overall SENTIMENT polarity of the INPUT sentence as Positive or Negative. \\ 
    & INPUT: <sent> \\
    & CLUES: <step-1-response>\\
    & \text{} \\
    \midrule
    {\bf \textit{Few-Shot}} & \\
    \specialrule{0em}{1pt}{1pt}
    \cdashline{1-2}
    \specialrule{0em}{1pt}{1pt}
    Classify Prompt: & Please classify the overall SENTIMENT polarity of the INPUT sentence as Positive or Negative. \\ 
    & INPUT: <demo-sent> \\
    & SENTIMENT: <demo-label-word> \\
    & \text{} \\
    & INPUT: <demo-sent> \\
    & SENTIMENT: <demo-label-word> \\
    & \text{} \\
    & INPUT: <sent> \\
    & SENTIMENT:  \\
    \specialrule{0em}{1pt}{1pt}
    \cdashline{1-2}
    \specialrule{0em}{1pt}{1pt}
    Reason-Classify Prompts: &  {\bf Step 1:}\\
    & Classify the sentiment of the input sentence as positive or negative. \\
    & INPUT: <demo-sent> \\
    & \text{} \\
    & {\bf Step 2:} \\ 
    & Classify the sentiment of the input sentence as positive or negative.  \\
    & \text{} \\
    & INPUT: <demo-sent> \\
    & REASONING: <step-1-generated>\\ 
    & SENTIMENT: <demo-label-word>\\
    & \text{} \\
    & INPUT: <demo-sent> \\
    & REASONING: <step-1-generated>\\ 
    & SENTIMENT: <demo-label-word>\\
    & \text{} \\
    & INPUT: <test-sent> \\
    \specialrule{0em}{1pt}{1pt}
    \cdashline{1-2}
    \specialrule{0em}{1pt}{1pt}
    Findclue-Reason-Classify Prompts: &  {\bf Step 1:}\\
    & Classify the sentiment of the input sentence as positive or negative. \\
    & INPUT: <demo-sent> \\
    & \text{} \\
    & {\bf Step 2:} \\ 
    & Classify the sentiment of the input sentence as positive or negative.  \\
    & \text{} \\
    & INPUT: <demo-sent> \\
    & REASONING: <step-1-generated>\\ 
    & SENTIMENT: <demo-label-word>\\
    & \text{} \\
    & INPUT: <demo-sent> \\
    & REASONING: <step-1-generated>\\ 
    & SENTIMENT: <demo-label-word>\\
    & \text{} \\
    & INPUT: <test-sent> \\
    \bottomrule
\end{tabular}
}
\caption{Examples of prompts for setups in Section~\ref{sec:methods}.}
\label{example:prompt_mr}
\end{table*}

\end{document}